# Markov Random Walk Representations with Continuous Distributions

**Chen-Hsiang Yeang**
MIT Artificial Intelligence Laboratory
200 Technology Square, Cambridge, MA 02139, USA

**Martin Szummer**
Microsoft Research
Cambridge CB3 0FB, U.K.

## Abstract

Representations based on random walks can exploit discrete data distributions for clustering and classification. We extend such representations from discrete to continuous distributions. Transition probabilities are now calculated using a diffusion equation with a diffusion coefficient that inversely depends on the data density. We relate this diffusion equation to a path integral and derive the corresponding path probability measure. The framework is useful for incorporating continuous data densities and prior knowledge.

## 1 Introduction

Many machine learning problems require evaluating pairwise relations (similarities or distances) between data points. Global Euclidean distances are often a poor measure of similarity when the data lies on a non-flat manifold. For example, the data may form a low-dimensional curved structure, embedded in a high-dimensional space, and similarity varies along that low-dimensional subspace. In such cases it is often better to define global distances by patching together local distances between neighboring points. Two examples of this approach are geodesic distances (Tenenbaum 1998) and Markov random walks (Tishby & Slonim, 2000).

Markov random walks provide a probabilistic model of similarity which is suitable for many classification and clustering tasks. The random walk defines a probability of transitioning between points. Transitions to a point are typically more likely to come from nearby points belonging to the same class or cluster than from points in other classes. Therefore, a point can be classified or clustered with points with high transition probabilities going to it.

Markov random walks are specified between discrete points. Firstly, one-step transition probabilities are defined from local distances (typically only nearby neighbors are included). The random walk then builds a global representation by making a fixed number of such local transitions, so that the representation follows the structure of the data. The number of transitions determines the scale of structures that are modeled. Specifically, after a large number of transitions, distinctions between nearby points will be lost, and the representation instead models coarse structures, e.g., big clusters of points.

Unfortunately, Markov random walks can only operate on given discrete data: all start-, end- and transition-points must be discrete. It is unclear how to incorporate a new data point without expensive recomputation of the whole representation. The discreteness is a fundamental limitation: frequently the discrete points are merely a finite sample from a continuous distribution. We would like to extend the framework to allow transitions via continuous space. There may also be an infinite number of start- and end-points, which again suggests a continuous formulation.

A continuous framework is needed for several application scenarios that contain continuous data. Occasionally, some of the data is directly provided as a continuous distribution. For example, we may have prior knowledge about the density of data in a particular region. Continuous data can result from translations from other representations: e.g. decision rules have been modeled as continuous regions of data (Fung et. al. 2002). Sometimes observations may be associated with complex measurement uncertainties, so that they are best modeled as distributions. Alternatively, invariances can be incorporated by letting each observation represent a whole set in space. Finally, it may be more computationally efficient to represent a large number of discrete points as a continuous distribution.

These examples highlight the importance of extending Markov random walks to a continuum of points.

Discrete random walks can be calculated using matrix powers. Unfortunately, this is no longer possible in the continuous case as the transition matrix would have to contain an uncountable number of entries. Instead, we take the limit as the number of data points grows, and arrive at a diffusion equation. Alternatively, we can derive the diffusion equation from the perspective of whole transition paths rather than single transitions. We assign probabilities to paths and estimate conditional probabilities by performing path integrals. The path integral formulation is shown to be equivalent to the diffusion equation.

Our approach is related to many previous works. Markov random walks on finite data have been applied in semi-supervised learning (Szummer & Jaakkola, 2001) and clustering (Tishby & Slonim, 2000). Recently, diffusion or heat kernels are applied in kernel-based machine learning algorithms (e.g., Kondor & Lafferty, 2002; Belkin & Niyogi, 2002, Lafferty & Lebannon, 2002). The link between path integrals and diffusion equations (Schrödinger's equations) was discovered during the early era of quantum mechanics and quantum field theory (Feynman, 1965). Several works in machine learning (e.g., Bialek, Callan & Strong, 1996; Nemenman & Bialek, 2000; Horn & Gottlieb, 2001) are also inspired by the mathematical techniques used in quantum theory.

The rest of the paper is organized as follows. Section 2 reviews Markov random walks and formulates a continuous version. Section 3 states the relation between the path integral formulation and the diffusion equation. Sections 4 to 6 contain experiments, discussion of potential machine learning applications that could benefit from continuous Markov random walks, and a conclusion.

## 2 Markov random walks

### 2.1 Discrete Markov random walks

We begin by reviewing the discrete Markov random walk representation. Suppose there are $m$ data points $\mathcal{B} = \{\mathbf{x}_0, \cdots, \mathbf{x}_{m-1} : \mathbf{x}_i \in R^n, \forall i\}$. We can construct a random walk on $\mathcal{B}$ and compute the conditional probability between each pair of points. Define a random variable $x$ whose states (values) are the points in $\mathcal{B}$. We can view this problem as a particle randomly transitioning among $m$ positions and $x(t)$ is the location of the particle at time $t$. Suppose initially we are certain that $x$ is in state $i$, i.e., the initial probability $Q_i^0(x) = (0, \cdots, 1, 0, \cdots, 0)^T$ whose $i^{th}$ entry is 1. The transition probability from $\mathbf{x}_i$ to $\mathbf{x}_j$ are inversely related to their Euclidean distance $d_{ij}$:

$$P_{ji} = \frac{1}{Z} e^{-\beta g(d_{ij})}, \quad (1)$$

where $g(.)$ is an increasing function. A point tends to transition to its neighbors according to equation 1. If $\mathcal{B}$ constitutes vertices of a graph then we set $P_{ji} = P_{ij} = 0$ whenever $(i,j)$ is not an edge of the graph. In our setting we may force the transition probabilities to vanish at non-neighbors in order to avoid distant jumps.

Given the initial state probability $Q_i^0(x)$ and state transition probabilities $P_{ji}$, it is straightforward to evaluate the state probability at time $t$:

$$Q_i^t(x) = (P)^t Q_i^0(x). \quad (2)$$

We express $P(.|.)$ as the conditional probability evaluated at time $t$.

$$\begin{aligned}P(\mathbf{x}_j, t | \mathbf{x}_i, 0) &\equiv P(x(t) = \mathbf{x}_j | x(0) = \mathbf{x}_i) \\ &= (Q_i^t(x))_j = (P)_{ji}^t.\end{aligned} \quad (3)$$

$P(\mathbf{x}_j, t | \mathbf{x}_i, 0)$ can be viewed as a similarity metric between $\mathbf{x}_i$ and $\mathbf{x}_j$. Different from global metrics such as Euclidean distances or radial basis kernels, it depends on both the locations of end points and the structure of the data manifold. If two points are close but located in a neighborhood where data points are sparse, then the conditional probability is low. The conditional probability also depends on $t$. If $t$ is long enough and $P_{ji}$ does not vanish, then all the points will mix up at $t$. The stationary distribution is an eigenvector of $P$. We want to stop at the stage where significant structures of the data (for example, clusters) are preserved.

### 2.2 Markov random walks in a continuum

A natural extension of discrete Markov random walks is to consider continuous datasets. Instead of a finite dataset $\mathcal{B} = \{\mathbf{x}_0, \cdots, \mathbf{x}_{m-1}\}$, we are given the density $\rho(\mathbf{x})$ over a Euclidean subspace. What will the mathematical form of the Markov random walk be in the continuous limit?

In the discrete case, we are interested in the conditional probability between each pair of data points in the forward ($P(\mathbf{x}(t) = \mathbf{x}_j | \mathbf{x}(0) = \mathbf{x}_i)$) or backward ($P(\mathbf{x}(0) = \mathbf{x}_i | \mathbf{x}(t) = \mathbf{x}_j)$) directions. In the continuous case, the interesting quantity is the transition density $P(\mathbf{x}(t) \in B_\epsilon(\mathbf{x}_j) | \mathbf{x}(0) \in B_\epsilon(\mathbf{x}_i))$, where $B_\epsilon(\mathbf{x})$ is the $\epsilon$-ball centered around $\mathbf{x}$. This density is the product of two quantities: the transition probability between two spatial locations, and the density of data points at the destination. The former can be calculated by extending the Markov random walk to a continuum of data points, and the latter is given. Here we distinguish between the transition densities of Markov random walks and the densities of data points.

Before directly tackling this problem, we first examine the continuous extension of a simpler problem. Suppose all data points are located at regular grids of equal distances. At each step, a particle is allowed to either stay at the current location or transition to one of its nearest neighbors. The number of nearest neighbors is $2d$, where $d$ is the dimension of the grid space. Assume the random walk is isotropic, i.e., the transition probability to each nearest neighbor is equal. This assumption seems to be inconsistent with the setting of the Markov random walk that transition probabilities are inversely related to pairwise distances (equation 1). However, as the grid size approaches to zero, the ratio $e^{-\beta d_1^2}/e^{-\beta d_2^2}$ should be very close to 1. Hence the probability of transitioning to each nearest neighbor is roughly equal.

Suppose a particle is initially located at $\mathbf{x} = 0$, we are interested in calculating the transition probability $P(\mathbf{x}, t)$ denoting the probability that the particle is located at $\mathbf{x}$ at time $t$ (given the initial condition that it is located at 0 at time 0). It can be shown that by reducing the grid size and the time interval to infinitesimal, $P(\mathbf{x}, t)$ follows the diffusion (heat) equation:

$$\frac{\partial P}{\partial t} = D\nabla^2 P, \qquad (4)$$

where $\nabla^2$ is the Laplacian operator, and the diffusion coefficient $D$ depends on the probability of staying at the current location, differential grid size and time interval. The solution of equation 2 is simply the Gaussian kernel

$$P(\mathbf{x}, t) = (\frac{1}{4\pi Dt})^{d/2} e^{-|\mathbf{x}|^2/4Dt}. \qquad (5)$$

If we push the time interval into the continuous limit but leave the data points to be discrete, then the solution becomes the heat kernel on graphs (Kondor and Lafferty, 2002). The Laplacian operator $\nabla^2$ is replaced by the Laplacian on a graph:

$$L_{ij} = \begin{cases} d_i & \text{if } i = j, \\ -1 & \text{if } i \sim j, \\ 0 & \text{otherwise.} \end{cases} \qquad (6)$$

where $d_i$ denotes the degree of vertex $i$ and $i \sim j$ denotes vertices $i$ and $j$ are adjacent. The graph Laplacian allows non-isotropic diffusion and transitions to non nearest neighbors. However, it cannot be directly extended into a continuous limit.

To derive the non-isotropic diffusion in a continuous limit, we consider a discrete example again. Instead of regular grids, we assume the grid sites are irregularly distributed in a $d$-dimensional Euclidean space. We can index each grid site by a $d$-tuple $I = (n_1, \cdots, n_d)$, where each $n_i$ is an integer denoting the index number along the $i$th coordinate. Except the grid sites at the boundary, each site $I$ has $2d$ nearest neighbors: $(n_1 \pm 1, \cdots, n_d), \cdots, (n_1, \cdots, n_d \pm 1)$. Unlike the regular grids, the distances between nearest sites can be different. Denote $\cdots, a_{i,-2}, a_{i,-1}, a_{i,1}, a_{i,2}, \cdots$ as the grid distances along the $i$th coordinate. Figure 1 illustrates an irregular grid on a 2D space.

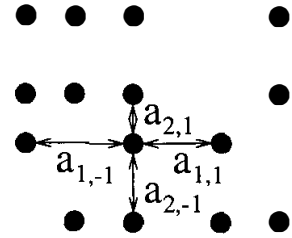

Figure 1: An irregular grid

We can depict a Markov random walk by the probability mass function $p_I(N)$ at each site $I$ and each time step $N$. Assume at each step, a particle stays at the current position with probability $p_0$, and hops to each nearest neighbor with equal probability $\frac{1-p_0}{2d}$. Then from a balance equation, the mass at site $I$ at time $N+1$ comes from either the same site or its neighbors at previous step $N$:

$$p_I(N+1) = p_0 p_I(N) + \frac{1-p_0}{2d}(\sum_{J \in Nd(I)} p_J(N)). \qquad (7)$$

Where $Nd(I)$ denotes nearest neighbors of site $I$. Thus,

$$\begin{aligned} p_I(N+1) - p_I(N) = \frac{1-p_0}{2d}\sum_{i=1}^{d}[(p_{I,n_i+1}(N) \\ -p_{I,n_i}(N)) - (p_{I,n_i}(N) - p_{I,n_i-1}(N))]. \end{aligned} \qquad (8)$$

$p_{I,n_i\pm 1}(N)$ denotes the probability mass at site $(n_1, \cdots, n_i \pm 1, \cdots, n_d)$.

To generalize the Markov random walk into a continuum, we reduce the time interval $\Delta$ and all grid sizes $\cdots, a_{i,-2}, a_{i,-1}, a_{i,1}, a_{i,2}, \cdots$ to infinitesimal values. The location $\mathbf{x}$ and time $t$ of a particle can be expressed as

$$\begin{aligned} \sum_{j=0}^{n_i} a_{i,j} &\to x_i. \\ N\Delta &\to t. \end{aligned} \qquad (9)$$

where $x_i$ is the $i$th component of $\mathbf{x}$. The probability mass function $p_I(N)$ reduces to the probability mass function $\phi(\mathbf{x}, t)$ at location $\mathbf{x}$ and time $t$:

$$p_I(N) \to \phi((\sum_{j=0}^{n_1} a_{1,j}, \cdots, \sum_{j=0}^{n_d} a_{d,j}), N\Delta) \to \phi(\mathbf{x}, t). \quad (10)$$

To derive the diffusion equation, we need to express equation 8 in a differential form. The left hand side is the time derivative of the probability mass function,

$$p_I(N+1) - p_I(N) \to \Delta \partial_t \phi(\mathbf{x}, t). \quad (11)$$

where $\partial_t$ denotes $\frac{\partial}{\partial t}$. The right hand side is the second order spatial derivative of the probability mass function,

$$\sum_{i=1}^{d} [(p_{I,n_i+1}(N) - p_{I,n_i}(N)) - (p_{I,n_i}(N) - p_{I,n_i-1}(N))] \to \sum_{i=1}^{d} a_{i,n_i}^2 \partial_{x_i}^2 \phi(\mathbf{x}, t). \quad (12)$$

where $\partial_{x_i}^2$ denotes $\frac{\partial^2}{\partial x_i^2}$.

Combining equations 11 and 12, equation 8 reduces to

$$\partial_t \phi(\mathbf{x}, t) = \sum_{i=1}^{d} D_i(\mathbf{x}) \partial_{x_i}^2 \phi(\mathbf{x}, t). \quad (13)$$

where

$$D_i(\mathbf{x}) = \frac{(1 - p_0) a_i^2(\mathbf{x})}{2d\Delta}. \quad (14)$$

Equation 13 is the continuous limit of the Markov random walk on irregular grids. The irregularity of the grids is naturally attributed from the non-uniform density of data points, since grid sites correspond to data points in the discrete case. Let $\rho(\mathbf{x})$ denote the density of data points at $\mathbf{x}$: there are $\rho(\mathbf{x}) \prod_i dx_i$ data points within a small hypercube centered at $\mathbf{x}$ with volume $\prod_i dx_i$. Within this small volume, assume grid lengths along all directions are equal, then the average grid length is

$$a(\mathbf{x}) = (\frac{\prod_i dx_i}{\rho(\mathbf{x}) \prod_i dx_i})^{\frac{1}{d}} = (\frac{1}{\rho(\mathbf{x})})^{\frac{1}{d}}. \quad (15)$$

Substituting 15 into 14, the diffusion coefficient has the same value along each direction.

$$D(\mathbf{x}) = \frac{(1 - p_0)}{2d\Delta \rho^{\frac{2}{d}}(\mathbf{x})}. \quad (16)$$

The diffusion equation becomes

$$\frac{\partial \phi(\mathbf{x}, t)}{\partial t} = D(\mathbf{x}) \nabla^2 \phi(\mathbf{x}, t). \quad (17)$$

Equation 17 is similar to the regular diffusion equation 4 except for a spatially varying diffusion coefficient. $D(\mathbf{x})$ in equation 16 is sensible in several aspects. First, the unit of $D(\mathbf{x})$ is $\frac{\text{length}^2}{\text{time}}$, which is consistent with the unit of the diffusion equation. Second, $D(\mathbf{x})$ decreases as $p_0$ increases, since the diffusion slows down if it spends more time staying at the same position. Third, $D(\mathbf{x})$ is inversely related to the density of data points, since the diffusion slows down as there are more sites to attract a particle. Fourth, $D(\mathbf{x})$ is inversely related to the dimension for the same reason, since there are more nearest neighbors in a higher dimensional space.

$\phi(\mathbf{x}, t)$ is the value of the probability mass function we would observe at each data point $\mathbf{x}$. However, when considering the true probability density function of the diffusion, we need to take the density of data points into account. Within a small volume, the number of data points is proportional to the density of data points. Thus the probability mass concentrated in this volume is proportional to $\phi(\mathbf{x}, t)$ times the density of data points:

$$\psi(\mathbf{x}, t) \propto \phi(\mathbf{x}, t) \rho(\mathbf{x}). \quad (18)$$

$\psi(\mathbf{x}, t)$ is the true probability density function of the diffusion.

## 3 Diffusion from path perspective

The diffusion processes can also be understood from the perspective of paths. This perspective is advantageous in continuous case, since transition probabilities depend on grid size and time intervals but path probabilities do not. If the formula of path probabilities is provided, then we can calculate their values without performing the random walks.

Suppose the initial and final positions of a particle are known, what is the probability that the diffusion process is realized along a specific path connecting the two points? If the Markov random walk is carried out on a discrete dataset, then the path probability can be explicitly evaluated once the transition probability is specified: it is simply the product of transition probabilities of edges along the path. When extending to a continuous dataset, there are uncountably many paths connecting the two end points. Conceptually, we can apply any probability measure on these paths as long as they integrate to 1. However, not all these measures are consistent with the diffusion equation 4 or 17. In this section, we demonstrate the existence of a path probability measure consistent with the diffusion equation. This measure is closely related to the regularization term in regressions and splines.

Consider an example of discrete random walks again. Let $f(t) = [\mathbf{x}_0, \mathbf{x}_{f(1)}, \cdots, \mathbf{x}_1]$ be a path connecting $\mathbf{x}_0$ at $t = 0$ and $\mathbf{x}_1$ at $t = T$. We choose the transition probability function in equation 1 as

$$P_{ji} = \frac{1}{Z} e^{-\beta d_{ij}^2}. \qquad (19)$$

Then the path probability of $f(t)$ is

$$\begin{aligned} P[f(t)] &\propto \prod_t \exp(-\beta d^2_{x_{f(t)}, x_{f(t+1)}}) \\ &= \exp(-\beta \sum_t d^2_{x_{f(t)}, x_{f(t+1)}}). \end{aligned} \qquad (20)$$

When extending to a continuous limit, each segment $d^2_{x_{f(t)}, x_{f(t+1)}}$ is proportional to the square derivative of $f(t)$, thus equation 20 becomes

$$P[f(t)] \propto \exp(\int_{f(t)} -\beta |\frac{df(t)}{dt}|^2 dt). \qquad (21)$$

Recall $\frac{df(t)}{dt}$ is a vector, and the integration is carried out along the path $f(t)$. $P[f(t)]$ can be rewritten as

$$P[f(t)] \propto \exp(-S[f(t)]), \qquad (22)$$

where the loss functional $S[f(t)]$ is

$$S[f(t)] = \int_{f(t)} \beta |\frac{df(t)}{dt}|^2 dt. \qquad (23)$$

The conditional probability from $(\mathbf{x}_0, t_0)$ to $(\mathbf{x}_1, t_1)$ is the integration of probabilities over all paths connecting these two points:

$$P(\mathbf{x}_1, t_1 | \mathbf{x}_0, t_0) = \int_{\mathbf{x}_0}^{\mathbf{x}_1} \frac{1}{Z} e^{-S[f(t)]} \mathcal{D}[f(t)]. \qquad (24)$$

Here $Z$ is a normalization constant, and $\mathcal{D}[f(t)]$ denotes the *path integral* over all functions $f(t)$ which satisfy boundary conditions $f(t_0) = \mathbf{x}_0$ and $f(t_1) = \mathbf{x}_1$. We can view the path integral as summing over an uncountable number of paths.

The loss functional in equation 23 is identical to the regularization term in regressions or splines. It penalizes the paths which fluctuate a lot over a short time interval ($|\frac{df(t)}{dt}|^2$ is large), so that smooth paths have larger path probabilities. However, our problem is different from regression/spline in two aspects. First, there is no data fitting term in our problem (e.g., the square error loss in regression). This is because the data points in this example are uniformly distributed. Second, unlike the non-Bayesian regression/spline whose goal is to find the path $f(t)$ that minimizes the loss functional, we are interested in the probabilities of all paths. The difference between finding the optimal path and evaluating the probabilities of all paths is analogous to finding the optimal trajectory in classical mechanics and computing the state transition propagator in quantum mechanics (Shankar, 1980).

The square differential loss function in equation 21 is interesting because the solution of its path integral problem in equation 24 exactly follows the diffusion equation 4. This equivalence was proven by Feynman (Feynman and Hibbs, 1965) during the early development of the quantum field theory.

**Theorem 1** Suppose $\mathcal{L} = \beta(\frac{dx}{dt})^2$, $P(x_1, t_1; x, t_0) = \int_{x_0}^{x_1} \frac{1}{Z} \exp\{-\int_{t_0}^{t_1} \mathcal{L} dt'\} \mathcal{D}[x(t)]$, and $P(x, t) = \int_{-\infty}^{\infty} P(x, t | x_0, t_0) P(x_0, t_0) dx_0$, then $P(x, t)$ follows the diffusion equation

$$\frac{\partial P(x, t)}{\partial t} = \frac{1}{4\beta} \frac{\partial^2 P(x, t)}{\partial x^2}. \qquad (25)$$

The proof of theorem 1 can be found in (Feynman, 1965; Shankar, 1980). A sketch of it is as follows. By discretizing time steps into $t_0, t_0 + \Delta t, \cdots, t_0 + N\Delta t = t_1$, the path integral $\int \mathcal{D}[x(t)]$ can be expressed as the multi-variable integral $\int \prod_i dx^i$, where $x^i = f(t_0 + i\Delta t)$ is the path value at time $t_0 + i\Delta t$. We consider the transition probability $P(x + \Delta x, t + \Delta t; x, t)$ between neighboring points. As $\Delta t \to 0$, we assume the straight line connecting the two points is dominant. The influence of all other paths vanishes due to the quadratic loss functional of the path derivatives. The path integral thus becomes an integral of Gaussian functions. By expressing it in a continuous limit, we obtain equation 25. Theorem 1 can be directly extended to multi-dimensional cases.

So far the density of data points has not been taken into account. The diffusion equation 4 and the path integral formulation 24 depict the random walks in a space where data points (grid sites) are uniformly distributed. To incorporate the density of data points into the picture, we re-examine equation 19. The transition probability depends not only on the distance between two points, but also on the density of the region between the two points. By comparing equations 17 and 25, the coefficient $\beta$ depends on the density of data points:

$$\beta(\mathbf{x}) = \frac{4}{D(\mathbf{x})}. \qquad (26)$$

$\beta(\mathbf{x})$ is inversely related to $D(\mathbf{x})$ and positively related to the data density $\rho(\mathbf{x})$. It imposes larger penalty on the paths which traverse the region of high data densities. This is because a path in a dense region experiences more transitions than a path of the same length in a sparse region. At a first glance, this property seems to contradict out intuition about path selection: a path which traverses in a dense region should be favored. However, since there are also more paths in a dense region, the aggregate probability that the transition is realized via paths in the dense region is still higher.

In many machine learning problems, data points are distributed on a low-dimensional manifold which is embedded in a high-dimensional Euclidean space. Recently, there are an increasing number of works of uncovering the underlying data manifold (e.g., Tenenbaum, 1998; Roweis & Saul, 2000; Belkin & Niyogim 2002) and constructing heat kernels on particular types of manifolds (e.g., Lafferty and Lebanon, 2002; Kondor and Lafferty, 2002). We can incorporate the information about the manifold in the diffusion coefficient and perform diffusion on the manifold. On a Riemannian manifold $M$, the tangent velocity along a path becomes

$$|\mathbf{v}(\mathbf{x},t)|^2 = \sum_{i,j} g_{ij}(\mathbf{x})(\frac{d\mathbf{x}^i}{dt})(\frac{d\mathbf{x}^j}{dt}), \quad (27)$$

where $g_{ij}$ is the *covariant metric tensor* of $M$. We use equation 21 to measure the unsmoothness of the path. Thus the loss functional of a curve $f(t)$ on $M$ becomes

$$S[f(t)] \propto \int \sum_{i,j} g_{ij}(f(t))(\frac{df^i}{dt})(\frac{df^j}{dt})dt. \quad (28)$$

$S[f(t)]$ is similar to equation 21 except the coefficient $g_{ij}$ is a vector. We can view it as having a different diffusion rate along each coordinate curve. Does the equivalence between the path integral formulation and the diffusion equation still hold? Theorem 2 extends the results of theorem 1 to a Riemannian manifold. Therefore, in principle we can compute the transition probabilities by solving diffusion equations on a manifold.

**Theorem 2**: Suppose $\mathcal{L} = g_{ij}(\mathbf{x})(\frac{d\mathbf{x}^i}{dt})(\frac{d\mathbf{x}^j}{dt})$, where $P(\mathbf{x},t;\mathbf{x}',0) = \int_{\mathbf{x}'}^{\mathbf{x}} \frac{1}{Z}\exp\{-\int_0^t \mathcal{L}dt'\}\mathcal{D}[\mathbf{x}(t)]$, and $P(\mathbf{x},t) = \int P(\mathbf{x},t;\mathbf{x}',0)P(\mathbf{x}',0)d\mathbf{x}'$, then $P(\mathbf{x},t)$ follows the diffusion equation

$$\frac{\partial P(\mathbf{x},t)}{\partial t} = Dg^{ij}(\mathbf{x})\frac{\partial^2 P(\mathbf{x},t)}{\partial \mathbf{x}_i \partial \mathbf{x}_j}, \quad (29)$$

for $\mathbf{x} \in M$, where $g^{ij}(\mathbf{x})$ is the contravariant metric tensor of $M$, which is the inverse matrix of $g_{ij}(\mathbf{x})$. The proof of theorem 2 is posted on our web page due to lack of space here.

The solution of equation 29 is the diffusion kernel on a Riemannian manifold. On certain types of manifolds, analytic solutions of diffusion kernels have been found. For instance, Lafferty and Lebanon (Lafferty & Lebannon, 2002) transformed data into parametric models and applied the diffusion kernels on the model manifold. Analytical solutions on multinomial and spherical Gaussian model manifolds were described in their paper.

## 4　Experiments

We first compared the outcomes of discrete and continuous random walks in one-dimensional space. We restricted the space of random walks in the interval $[-1, 1]$. The density of data points in the region $[0, 1]$ is ten times of the density in the region $[-1, 0]$. In discrete experiments, we place data points regularly with distance 0.01 on $[-1, 0]$ and distance 0.001 on $[0, 1]$. In continuous experiments, we set the diffusion coefficient on $[-1, 0]$ to be 100 times the diffusion coefficient on $[0, 1]$ according to equation 16. The absolute scale of diffusion coefficients is not critical since its effect can be compensated by adjusting the stopping time of diffusion.

For discrete random walks, we set the self-transition probability to 0.2. All internal points have left and right neighbors which are assigned equal transition probabilties. The two edge points can only transition to one neighbor (in addition to a self-transition). The initial distribution is set to one for a single point and zero elsewhere. Then the probability mass function is updated according to equation 3.

For continuous random walks, we applied the finite difference method to solve the diffusion equation 17. The spatial grid size is 0.001, the temporal step interval is 0.1, and the diffusion coefficients to the right and left are $5 \times 10^{-9}$ and $5 \times 10^{-7}$. These small values are chosen in order to avoid overshoots of update values. Initially the probability mass is a delta function concentrated at one point. The (continuous) probability mass function is updated by the following finite difference equation:

$$\phi(x, t+\Delta t) = \phi(x,t) + D(x)\frac{\Delta t}{(\Delta x)^2}(\phi(x+\Delta x, t)+ \phi(x-\Delta x,t) - 2\phi(x,t)). \quad (30)$$

For appropriate choices of parameters, the discrete and continuous probability mass functions should have the same shape. We set the diffusion constant inversely proportional to the square of the data density as indicated by equation 16. Figure 2 demonstrates that continuous and discrete random walks lead to very similar probability distributions. The probability mass is initialized to be 1 at $x = 0$. In both cases, the mass diffuses faster in the sparse region $x < 0$. Moreover, the shapes of the probability mass function are very similar in these two cases. This suggests that the difference of diffusion coefficients in the continuous case is consistent with the data density in the discrete case.

Figure 3 shows the random walk outcomes with the mass initialized to 1 at $x = -0.1$. The stopping times are identical to the previous experiment. The diffusion proceeds faster to the left but reaches a bottleneck at $x = 0$. The probability mass functions of discrete and continuous cases are slightly different: the discrete case is not smooth at $x = 0$, but would become smoother if we reduced the interval between data points. The continuous case is perfectly smooth since it reflects infinitesimal intervals.

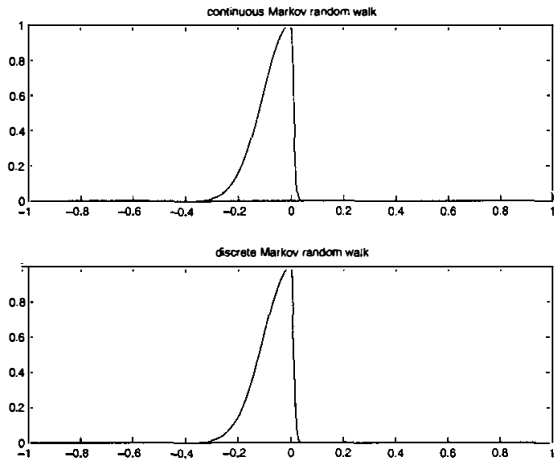

Figure 2: Comparisons of continuous and discrete random walks, starting from $x = 0$. The peak has been scaled to 1.

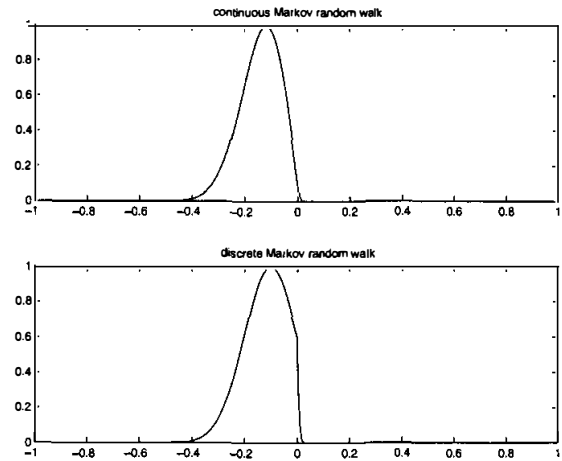

Figure 3: Comparisons of continuous and discrete random walks, starting from $x = -0.1$.

### 4.1 Swiss-roll data experiment

Suppose we have prior information that the data is distributed continuously as in figure 4. There are two narrow high-density regions embedded in space which has 200 times lower data density. For machine learning tasks, we would like to construct a similarity measure that respects the data density. For discrete data, transition probabilities given by Markov random walks can be used for this purpose. Here, we construct transition probabilities by diffusion.

Again we solve the diffusion equation numerically via the finite difference method. The space was discretized to a grid of 90 by 110 nodes, and the solution was assumed to decay exponentially at the boundaries. The time was discretized uniformly in the range of 0 to 20000[1].

The diffusion was initialized with an impulse. Initially, the diffusion follows shortest paths, and the impulse spreads radially as shown for time $t=3333$ (Figure 5, left). However, as time passes, the transition probabilities are no longer radial, and instead the diffusion is dominated by paths following the high-density re-gion (at $t=20000$). The transition probability also approaches a constant in a large region around the starting point. However, the transition probability to the low-density region is small everywhere.

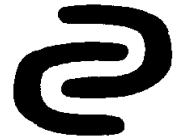

Figure 4: Continuous swiss-roll data. The black regions have 200 times higher data density than the white backround.

## 5 Discussion

Continuous Markov random walk representations can be used in the same contexts as their discrete counterparts. For example, the probability of transitioning $P(\mathbf{x}_i, 0 | \mathbf{x}_j, t)$ can be used inside a classifier of form $P(y|\mathbf{x}_j) = \sum_i P(y|\mathbf{x}_i)P(\mathbf{x}_i, 0|\mathbf{x}_j, t)$, where $P(y|\mathbf{x}_i)$ are weights to be trained. This classifier only considers similarities between a finite set of discrete points; while the diffusion provides similarities to all points in space, we can select a finite set of interest by conditioning on them. This scheme still allows us to benefit from continuous information that was present during the diffusion, potentially giving rise to a more accurate similarity measure. In a classifier based on a discrete Markov random walk, only the given points $\{\mathbf{x}_i\}$ would be used, whereas here we may have incorporated prior knowledge, invariances, or known unlabeled data distributions.

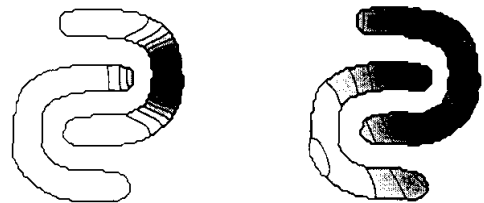

Figure 5: Diffusions starting from the middle of the right arm of the swiss roll, at time $t=3333$ (left), and $t=20000$ (right). There is a high transition probability to the dark areas.

---

[1] Time increments were $\Delta=0.1$, diffusion constants $D(x)=1.5 \times 10^{-5}$ in the high-density and $3 \times 10^{-3}$ in the low-density regions.

Continuous Markov random walks give similarity measures to any point in space. This ability is sorely missing from discrete Markov random walks, which define transition probabilities only to a priori given points in the space, restricting them to transductive classification. In fact, the continuous representation is also a valid kernel for a kernel classifier such as an SVM:

$$f(\mathbf{x}) = \sum_{i=1}^{n} \alpha_i y_i K(\mathbf{x}, \mathbf{x}_i). \quad (31)$$

This is valid because the heat kernel is a positive semi-definite kernel in Euclidean and Riemannian spaces (Itô, 1991)). Hence we can merely substitute the transition probability density $P(\mathbf{x}, t; \mathbf{x}_i, 0)$ for the kernel function. Again, such a kernel function can reflect the data distribution on the manifold, and experimentally tends to provide good classification results in the discrete case (Szummer & Jaakkola, 2001).

Continuous Markov random walk representations do face some challenges in practice. First, continuous Markov random walks are typically more computationally demanding than discrete walks. To compute the conditional probability density from each point, we need to solve the diffusion equation with an impulse initial condition. Solving differential equations is certainly more involved than matrix multiplications. Second, numerical differential equation solvers cannot directly handle high data dimensions. Finite difference methods discretize space into small volumes, whose numbers grow exponentially with the dimension of the embedding space. One approach to get around this curse of dimensionality is to restrict the support of the density function to a low-dimensional subspace. A final challenge remains to determine the period of diffusion, which amounts to determining the smoothness of the data.

## 6 Conclusion

In this paper, we have proposed a framework for extending discrete Markov random walks to continuous data distributions. Transition probabilities are now calculated using a diffusion equation. We have related this diffusion equation to a path integral and derived the corresponding path probability measure.

Our method provides a way to extend Markov random walks to incorporate both observed data and prior information about data distribution.